\pdfoutput=1

\documentclass[11pt]{article}

\usepackage[final]{acl}

\usepackage{times}
\usepackage{latexsym}

\usepackage[T1]{fontenc}

\usepackage[utf8]{inputenc}

\usepackage{microtype}

\usepackage{inconsolata}

\usepackage{graphicx}

\usepackage[utf8]{inputenc} 
\usepackage[T1]{fontenc}    
\usepackage{hyperref}       
\usepackage{url}            
\usepackage{booktabs}       
\usepackage{amsfonts}       
\usepackage{nicefrac}       
\usepackage{microtype}      
\usepackage{graphicx}

\usepackage{tabularx}
\usepackage{booktabs}
\usepackage{multirow}
\usepackage{makecell}

\usepackage{tcolorbox}
\usepackage{mdframed}

\usepackage{listings}
\usepackage{color}

\usepackage{amsmath}

\title{A Calibrated Reflection Approach for Enhancing Confidence Estimation in LLMs}

\author{Umesh Bodhwani, Yuan Ling, Shujing Dong, Yarong Feng, Hongfei Li, Ayush Goyal \\
Amazon.com \\
Seattle, Washington, USA \\
\texttt{\{bodhwani, yualing, shujdong, yarongf, lihongfe, ayushg\}@amazon.com} \\
}

\begin{document}
\maketitle

\begin{abstract}

A critical challenge in deploying Large Language Models (LLMs) is developing reliable mechanisms to estimate their confidence, enabling systems to determine when to trust model outputs versus seek human intervention. We present a Calibrated Reflection approach for enhancing confidence estimation in LLMs, a framework that combines structured reasoning with distance-aware calibration technique. Our approach introduces three key innovations: (1) a Maximum Confidence Selection (MCS) method that comprehensively evaluates confidence across all possible labels, (2) a reflection-based prompting mechanism that enhances reasoning reliability, and (3) a distance-aware calibration technique that accounts for ordinal relationships between labels. We evaluate our framework on diverse datasets, including HelpSteer2, Llama T-REx, and a proprietary conversational dataset, demonstrating its effectiveness across both conversational and fact-based classification tasks. This work contributes to the broader goal of developing reliable and well-calibrated confidence estimation methods for LLMs, enabling informed decisions about model trust and human judgement.

\end{abstract}

\section{Introduction}

LLMs have revolutionized many domains, but ensuring their outputs are trustworthy remains a pressing challenge. A key aspect of this trustworthiness is confidence estimation—developing methods to gauge the likelihood of an LLM's answer being correct. This is challenging due to the frequent miscalibration of their confidence scores. In traditional classification, a model's predicted probability can serve as a confidence estimate, but these probabilities must be well-calibrated to be meaningful. Calibration ensures that if a model claims $90\%$ confidence, it should be correct about $90\%$ of the time. In practice, an LLM might generate a fluent, plausible-sounding answer with near-certain confidence, yet be factually wrong - an undesirable situation if not detected by a confidence calibration mechanism. Techniques like chain-of-thought reasoning and self-consistency have been explored to improve the model’s self-evaluation, yet often yield overconfident estimates.

The challenge of confidence estimation becomes more nuanced in ordinal classification problems (e.g., user ratings, sentiment levels, risk assessments). Unlike nominal categories, ordinal labels enable consideration of distance between predictions: mistaking a rating of $5$ for $4$ is a smaller error than mistaking it for $1$. However, most existing confidence estimation methods treat each label independently, failing to differentiate between \textit{close} and \textit{far} errors. This limitation is particularly critical in sensitive settings where miscalibrated confidence on an ordinal decision can have serious consequences.

Existing approaches to confidence estimation in LLMs can be broadly categorized into four categories: (1) probability-based methods that utilize model logits and calibration algorithms~\citep{guo2017calibration}, but are restricted by the availability of model logits. (2) Fine-tuning methods require extensive training data and, while effective on in-domain datasets, struggle with generalization to out-of-domain scenarios. (3) Prompting-based techniques elicit self-evaluation from the model \cite{kadavath2022language}. Despite their intuitiveness appeal, these methods frequently yield overconfident estimates, undermining their reliability. (4) Ensemble methods~\citep{wang2022self, wang2024multi} can enhance reliability, but they incur significant computational overhead and do not inherently address the fundamental issue of calibrating confidence scores.

A critical gap in current research is the lack of methods that account for ordinal relationships between labels. Recent work \citep{qin2024lampo} suggests that effective confidence estimation requires both robust reasoning capabilities and proper calibration of confidence scores.

In this paper, we propose a novel framework for confidence estimation that addresses these limitations through a synergistic combination of advanced prompting techniques and distance-aware calibration. Our approach integrates:
\begin{itemize}
\item \textbf{Maximum Confidence Selection (MCS)} method that comprehensively evaluates confidence across all possible labels
\item \textbf{Reflection-based prompting} that enhances the model's reasoning to yield more reliable confidence estimates
\item \textbf{Distance-aware calibration} technique that accounts for ordinal relationships among labels
\end{itemize}

We evaluate the Calibrated Reflection approach on diverse datasets, including conversational and fact-based classification tasks. Across multiple benchmarks, our framework consistently improves confidence calibration and overall predictive performance. We observe significantly lower Expected Calibration Error (ECE) and Brier Score (indicating better alignment between predicted confidence and actual accuracy) compared to baselines, while also achieving higher discrimination metrics like AUROC and AUPRC. These improvements hold without any fine-tuning of the LLM and making a \textbf{single LLM invokation}, making our framework readily applicable in real-world scenarios. Our approach, integrating structured reasoning with distance-aware calibration, significantly advances confidence estimation for LLMs, enabling well-calibrated confidence scores in ordinal classification, thereby enhancing reliability and trustworthiness in LLM-driven applications.

\begin{figure*}[t]
    \centering
    \includegraphics[width=0.9\linewidth]{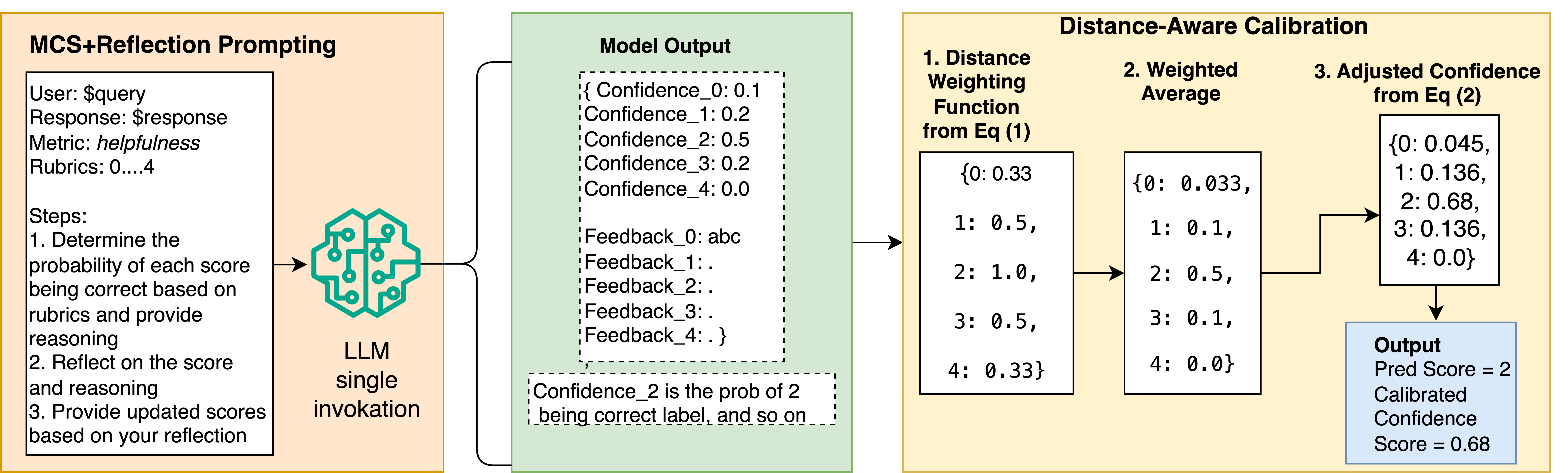} 
    
    \caption{Calibrated Reflection Workflow: Reflection Prompting generates an initial confidence score based on rubrics, reflects on its reasoning, and updates the scores.  Distance-Aware Calibration adjusts the scores based on ordinal distances between labels. The output includes the predicted score and the calibrated confidence score}
    \label{fig:confidence_score_estimation}
\end{figure*}

\section{Related Work}

Existing approaches to confidence estimation in LLMs have evolved from basic probability-based methods to more sophisticated techniques incorporating multiple strategies. 
\textbf{1)} Early methods~\citep{jiang2020can} focused on \textbf{sequence probability}, which estimates confidence by computing average log probabilities assigned to output tokens. While these approaches have been applied in various contexts, including close tasks and QA setups~\citep{muhlgay2023generating}, they require well-calibrated probabilities to accurately reflect correctness~\citep{guo2017calibration}, and generally don't represent the actual probability of the predicted results in LLMs.
\textbf{2) Verbalized confidence estimation} has emerged as a direct approach where LLMs assess their own confidence~\citep{kadavath2022language}. Chain of Thought prompting~\citep{wei2022chain} improves explanation and justification by breaking down reasoning into smaller steps. Self-consistency~\citep{wang2022self} estimates confidence by evaluating consensus across multiple reasoning paths. Recent extensions have incorporated debate-style prompting~\citep{irving2018ai} and reflection prompting~\citep{shinn2024reflexion}, where models internally challenge their decisions and self-assess potential errors.
\textbf{3) Model aggregation methods:} Combine signals from multiple sources for confidence estimation. While ensemble methods~\citep{zhang2020mix} merge outputs from multiple LLMs at high computational cost, the ReScorer~\citep{mohta2024rescorer} offers a more efficient approach by aggregating multiple ROSCOE metrics into comprehensive confidence scores.
\textbf{4) }\textbf{Surrogate models}~\citep{shrivastava2023llamas} have been proposed to assess main model outputs, with extensions like MPC~\citep{yang2024verbalized} incorporating knowledge injection from stronger models. The trained probe method~\citep{mahaut2024factual} represents a newer approach, training lightweight models on LLM internal representations to extract confidence signals. Uncertainty-aware Instruction Tuning (UaIT)~\citep{liu2024can} presents a promising direction in self-training, aligning LLMs' uncertainty perception with their outputs.

\section{Methodology}

\subsection{Problem Definition}
Given a LLM $M$ and an input sequence $X$, let $Y = M(X)$ denote the model-generated output. We aim to develop a confidence estimation framework that predicts the reliability of the model's output. Formally, we define a confidence function $C$ that maps the model's output to a confidence score:
\begin{equation}
    C(Y) \rightarrow [0,1]
\end{equation}
A confidence score close to 1 indicates high confidence in the output's reliability, while a score close to 0 indicates low confidence. The objective is to ensure that:
\begin{equation}
P(R(Y) = 1 \mid C(Y) = p) \approx p
\label{eq:formulation}
\end{equation}
where $R(Y)$ is a binary function indicating whether the output $Y$ is correct ($1$) or incorrect ($0$), and $p$ is the predicted confidence level. 

This formulation captures the essential goal of developing a well-calibrated confidence estimation system, ensuring that confidence scores align with the actual likelihood of correctness.

\begin{figure*}[t]
\small
\begin{tcolorbox}[title=Prompt for MCS-R method]
\begin{verbatim}
You are an AI judge tasked with the assessment of the quality of interaction between a user and a
conversation agent. You are presented with a single-turn interaction between the USER and AGENT,
which contains a USER utterance/request and a conversational AGENT response.

### Metric: {metric}
### Score Rubrics: {rubrics}

### Your tasks:
1. For each label, provide your **initial feedback** on whether the given label is correct.
2. Reflect on your reasoning to identify any potential errors or oversights.
3. Provide your **final feedback** after reflection.
4. Estimate the probability (between 0 and 1) that the given label is correct.

### Please output the following:
**feedback_n:** Summary of your initial evaluation, any adjustments or insights after reflection,
and your final evaluation of nth score in Score Rubrics
**score_n:** Probability of nth score in Score Rubrics being correct
---
USER: {user_query}
AGENT: {agent_response}
\end{verbatim}
\end{tcolorbox}
\caption{MCS-R prompt template for conversation quality assessment. The model follows a structured evaluation process (initial feedback → reflection → final feedback → probability) for each potential label}

\label{fig:prompt_template}
\end{figure*}

\subsection{Calibrated Reflection Approach}

To meet the calibration objective in Eq.~\ref{eq:formulation}, we propose a two-component framework comprising: (1) a confidence elicitation mechanism using Maximum Confidence Selection (MCS) with reflection-based prompting, and (2) a distance-aware calibration procedure. The first component obtains a comprehensive distribution of confidence scores across all candidate labels, enhanced by a reflection step to improve reasoning reliability. The second component then adjusts and calibrates the selected confidence score by accounting for the ordinal relationships between labels. Together, these components produce a well-calibrated confidence estimate for the model’s output, particularly effective in ordinal classification tasks.

\subsubsection{Eliciting Confidence through MCS and Advanced Prompting}
Prior prompting-based methods~\cite{tian2023justaskcalibrationstrategies} often focus on a limited set of top-$k$ most likely options, which can miss information about the model’s uncertainty over the full label space. We extend this to a Maximum Confidence Selection (MCS) approach that evaluates all labels. Formally, let $\mathcal{L}={y_1, y_2, \dots, y_n}$ be the set of all possible labels for the task. Given an input $x$, we prompt the model to assign a confidence score $C(x,y_i)$ to each label $y_i$, which denotes the model’s estimated probability that $y_i$ is the correct label for $x$.

We implement reflection-based prompting to elicit probability estimates.
The prompt shown in Figure~\ref{fig:prompt_template} first presents the context and the set of candidate labels (along with any task-specific definitions or rubrics) and then instructs the model to go through an evaluate–reflect–conclude process for each label. This structured prompting draws inspiration from self-reflection~\citep{selfreflection} and chain-of-thought~\citep{wei2022chain} techniques, encouraging the model to internally verify its initial answers before committing to a probability. Empirically, this approach, which we term MCS-R (Multiple Choice Scoring with Reflection) demonstrates improved calibration through reduced overconfidence and enhanced reasoning consistency compared to prior~\citep{Mahaut_2024, tian2023justaskcalibrationstrategies} approaches.

\subsubsection{Distance-Aware Calibration}

While MCS-R yields a probability distribution over labels, we further calibrate the model's overall confidence by considering the structure of the label space. In tasks with ordinal labels, not all errors are equally severe: predictions closer to the correct label should inspire more confidence than distant ones. We introduce a distance-aware calibration technique to adjust the confidence of the predicted label $\hat{y}$ based on how the remaining probability mass is distributed across labels near to vs. far from $\hat{y}$ in the label ordering. This approach builds on the insight that well-calibrated probabilities should reflect the model's uncertainty smoothly across adjacent labels and aligns with established calibration methods for probabilistic models.

Let $E$ be the index of the predicted label $\hat{y}$ in the ordered label set (for example, if $\hat{y}=4$ on a 5-point scale, then $E=4$). For each label index $i$, we define a distance-based weight that decreases as $i$ is farther from $E$:
\begin{equation}
    W(i,E) = \frac{1}{1 + |i-E|}
\end{equation}

Here $W(i,E)=1$ when $i=E$ (the predicted label), $W(i,E)=1/2$ for labels one step away, $W(i,E)=1/3$ for labels two steps away, and so on. This weighting function encodes ordinal relationships between labels, assigning larger weights to labels closer to the predicted class. Intuitively, $W(i,E)$ measures how confidence in label $i$ influences confidence in label $E$: high probability for nearby labels (small $|i-E|$) is less concerning than for distant labels.

Using these weights, we compute an adjusted confidence for the label $E$ as a weighted aggregate of the model’s original confidence scores $C_i$:
\begin{equation}
    \textit{Adjusted Confidence}_E = \frac{\sum_{i \in \mathcal{L}} C_i \cdot W(i,E)}{\sum_{i \in \mathcal{L}} C_i + \epsilon}
\end{equation}
where $\mathcal{L}$ is the set of all label indices. The denominator ensures the final confidence lies in the range $[0,1]$, and $\epsilon$ is a small positive constant (e.g., $10^{-6}$) to prevent division by zero in extreme cases where all confidence scores are zero. This formulation produces an adjusted confidence score that accounts for both the magnitude and distribution of the model's confidence across the ordinal label space. High probabilities assigned to labels far from $\hat{y}$ reduce the adjusted confidence, reflecting increased prediction uncertainty.

\section{Experimentation}
In this section, we outline the datasets, evaluation metrics, and comparison methods. We experiment with Claude-3-Haiku~\citep{claude3haiku2024} (closed-source), and Mistral-7B-instruct (open-source) models~\citep{jiang2023mistral}, and perform all experiments in a zero-shot setting, utilizing a fixed temperature of $0.1$, with single LLM invokation.

\subsection{Datasets}
\label{sec:datasets}
We evaluate our framework on three datasets: two conversational datasets and one fact-based classification dataset.

\textbf{HelpSteer2 (Conversational)~\citep{wang2024helpsteer2}:} Benchmark dataset designed to evaluate LLM-generated responses across five dimensions, \textit{helpfulness}, \textit{correctness}, \textit{coherence}, \textit{complexity}, \textit{verbosity}. The evaluation dataset consists of 1038 single-turn conversations, annotated with ordinal labels ranging from $0$ to $4$ for all metrics.

\textbf{Llama T-REx (Fact-Based Classification)~\citep{elsahar2018t}:} Following~\citep{Mahaut_2024}, we construct an evaluation dataset of $13.6K$ examples, comprising $6.8K$ true statements paired with their corresponding false counterparts.

\begin{figure*}[t]
    \centering
    \includegraphics[width=0.9\linewidth]{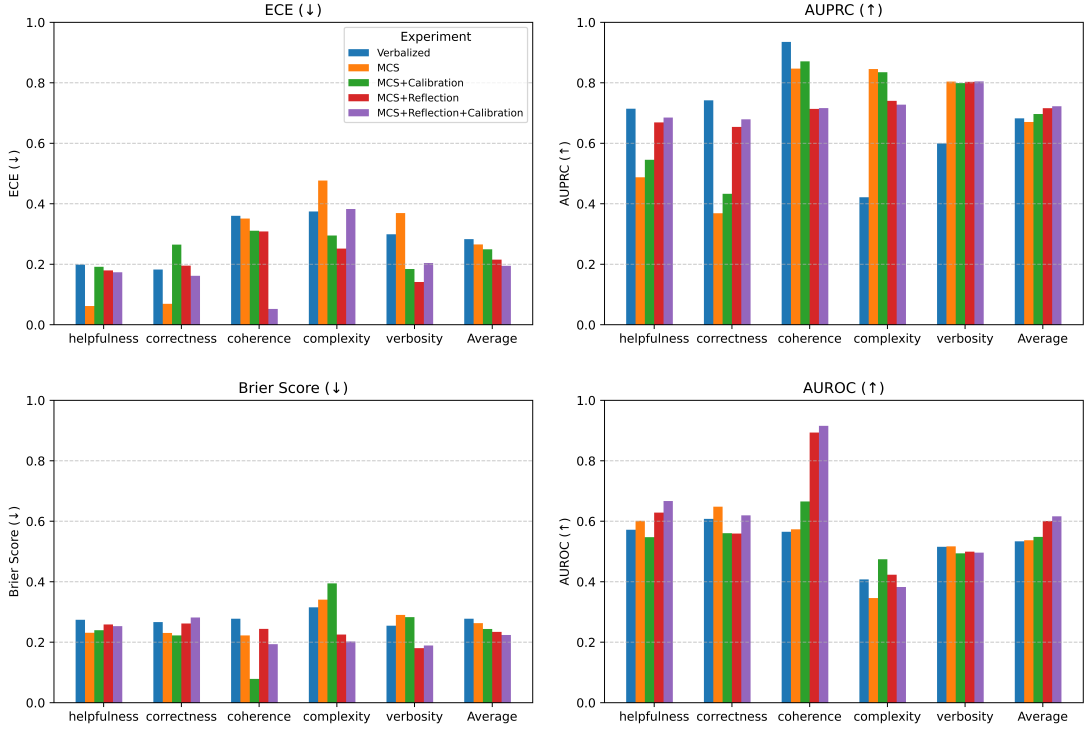} 
    
    \caption{Performance comparison on the HelpSteer2 dataset across different evaluation metrics. Results are reported for VC, MCS, MCS-C, MCS-R, and MCS-RC. Each bar group represents performance on five conversational dimensions and their average, highlighting the impact of advanced prompting techniques and calibration on confidence estimation.}
    \label{fig:helpsteer}
\end{figure*}

\textbf{Conversational Dataset:} A proprietary dataset of $314$ multi-turn conversations, each averaging six turns. With eight conversational dimensions (\textit{Issues}, \textit{Friction}, \textit{Task Success}, \textit{Info Factuality}, \textit{Coherence}, \textit{Naturalness}, \textit{Comprehensiveness}, \textit{Length}), this yields approx $15K$ evaluation points. Each turn is annotated by two independent contractors, with a third reviewer resolving discrepancies. This dataset evaluates our framework's effectiveness in real-world, multi-turn conversational settings.

\subsection{Evaluation Metrics}
We evaluate our methods using four complementary metrics: AUPRC (precision-recall trade-off for imbalanced datasets), AUROC (discriminative ability via true/false positive rates), ECE (calibration quality through confidence-correctness alignment), and Brier Score (overall calibration and accuracy via mean squared error). While AUPRC and AUROC assess discriminative performance, ECE and Brier Score measure calibration quality. Detailed metric calculations are provided in Appendix~\ref{app:eval_metrics}.

\subsection{Compared Methods}
\subsubsection{Baselines}
\textbf{Verbalized Confidence (VC):} Following \citep{tian2023justaskcalibrationstrategies}, this method prompts the model to output a confidence score (0-1) after each answer.

\textbf{Trained Probe (TP):} These methods transform LLM's internal representations from final or earlier layers into confidence scores, leveraging learned patterns for task-specific calibration.

\textbf{Log Probability (LP):} This approach averages token-level log probabilities of the output sequence to estimate confidence, building on established calibration work~\citep{guo2017calibration,xiong2023can}.

\textbf{Self-Consistency (SC):} This method generates multiple answers and computes confidence based on answer agreement rate, following~\citep{wang2022self}. Higher agreement among independent generations indicates higher confidence.

\textbf{Top-K Confidence (TK):} Drawing from~\citep{tian2023justaskcalibrationstrategies}, this approach prompts the model to elicit confidence for $top-k(k=2,4)$ predictions.

\subsubsection{Proposed Methods}

\textbf{Maximum Confidence Selection (MCS):} Our base method computes confidence scores for all labels, selecting the highest-scoring label as the prediction. Unlike conventional top-k approaches, MCS evaluates the complete label set for comprehensive confidence distribution.

\textbf{MCS+Reflection (MCS-R):} Enhances MCS with reflection-based prompting, where the model evaluates each label's plausibility, then refines its reasoning through reflection before assigning final confidence scores. This process improves estimation robustness and interpretability.

\textbf{MCS+Calibration (MCS-C):} Incorporates distance-aware calibration using a weighting function that accounts for ordinal relationships between labels. This adjustment prioritizes scores closer to the predicted label, reducing overconfidence.

\textbf{MCS+Reflection+Calibration (MCS-RC):} Combines reflection-based prompting with distance-aware calibration to create a comprehensive framework. Reflection provides structured reasoning while calibration ensures ordinal alignment. Prompt templates are provided in Appendix~\ref{app:prompt_templates}.

\begin{table}[t]
\centering
\caption{Average Scores for Confidence Score Estimation Experiments on Verbalized Method}
\begin{tabular}{@{\hspace{2pt}}l@{\hspace{2pt}}|@{\hspace{2pt}}c@{\hspace{2pt}}|@{\hspace{2pt}}c@{\hspace{2pt}}|@{\hspace{1pt}}c@{\hspace{1pt}}|@{\hspace{1pt}}c@{\hspace{1pt}}}
\hline
\textbf{Method} & \textbf{Brier} & \textbf{ECE} & \textbf{AUROC} & \textbf{AUPRC} \\
\hline
Log Prob & 0.117 & 0.101 & 0.623 & 0.895 \\
CoT & \underline{0.098} & 0.093 & 0.667 & 0.908 \\
Self-Consistency & \textbf{0.095} & 0.086 & 0.672 & 0.913 \\
Debate & 0.104 & \textbf{0.078} & \textbf{0.692} & \textbf{0.927} \\
Reflection & 0.110 & \underline{0.080} & \underline{0.687} & \underline{0.918} \\
\hline
\end{tabular}
\label{tab:verbalized_rd}
\end{table}

\begin{table}[t]
\centering
\caption{Comparison of Results for MCS with Reflection and Debate Prompts}
\begin{tabular}{l|c|c|@{\hspace{2pt}}c@{\hspace{2pt}}|@{\hspace{2pt}}c@{\hspace{2pt}}}
\hline
\textbf{Method} & \textbf{Brier} & \textbf{ECE} & \textbf{AUROC} & \textbf{AUPRC} \\
\hline
MCS & .138 & .169 & .591 & .899 \\
+Reflection & .256 & .301 & .697 & .925 \\
+Debate & .256 & .322 & .700 & .924 \\
\hline
\end{tabular}
\label{tab:mcs_reflection_debate}
\end{table}

\section{Results and Discussion}

\subsection{Effectiveness of Calibrated Reflection}

We evaluate the proposed confidence estimation framework on HelpSteer2 dataset using five methods: VC, MCS, MCS-R, MCS-C, and MCS-RC. Results, presented in Figure~\ref{fig:helpsteer}, include four evaluation metrics: ECE and Brier Score (lower is better), as well as AUPRC and AUROC (higher is better), computed across five conversational dimensions: \textit{helpfulness, correctness, coherence, complexity, and verbosity}, along with their average. On average, MCS-RC achieves the best performance across all metrics, validating the effectiveness of combining reflection-based reasoning with calibration. Notably, while VC slightly outperforms MCS on AUPRC, MCS exhibits superior performance on ECE, Brier Score, and AUROC, indicating that its comprehensive consideration of all labels enhances overall calibration and discriminative ability. MCS-R significantly improves performance compared to MCS, highlighting its ability to refine confidence estimates through iterative feedback, while MCS-C further enhances calibration quality by accounting for ordinal relationships, albeit with a smaller impact. Combining both techniques (MCS-RC) yields consistently superior results across all metrics, demonstrating their complementary nature. Metric-wise, MCS-RC excels in helpfulness and correctness, achieving the lowest calibration errors and highest discriminative scores, while MCS-R dominates coherence. For complexity and verbosity, calibration plays a more prominent role, effectively leveraging ordinal relationships. These findings validate MCS-RC as a robust method for confidence estimation, demonstrating superior calibration and discriminative performance across diverse conversational dimensions.

\subsection{Performance on Real-World Dataset}
To evaluate the robustness of our framework, we conduct experiments on a proprietary conversational dataset, systematically analyzing the effects of advanced prompting strategies and calibration. These experiments are divided into three key stages: \textbf{verbalized prompting, MCS}, and \textbf{MCS with calibration and enhanced prompts}. Verbalized prompting is conducted using Chain of Thoughts, Self-Consistency, Debate, and Reflection Prompting. Details about these methods are in Appendix \ref{app:verbalized_methods}. The results are summarized in Tables \ref{tab:verbalized_rd}, \ref{tab:mcs_reflection_debate}, and \ref{tab:mcs_ref_calib}, and detailed findings are presented below.

\begin{table*}[t]
\centering
\caption{Final Evaluation Scores Combining MCS + Enhanced Prompts + Calibration Technique}
\begin{tabular}{lcccccc}
\hline
\textbf{Method} & \textbf{Brier Score} & \textbf{ECE} & \textbf{AUROC} & \textbf{AUPRC} \\
\hline
MCS & 0.13799 & \textbf{}{0.16877} & 0.59146 & 0.89905 \\
MCS+Debate (Ours) & 0.25574 & 0.32192 & 0.69999 & 0.9243 \\
MCS+Reflection (Ours) & 0.25617 & 0.30092 & 0.6969 & 0.92534 \\
MCS+Debate+Calibration (Proposed) & \underline{0.1282} & 0.19243 & \underline{0.73752} & \underline{0.93334} \\
MCS+Reflection+Calibration (Proposed Best) & \textbf{0.12502} & \underline{0.17472} & \textbf{0.73994} & \textbf{0.93516} \\
\hline
\end{tabular}
\label{tab:mcs_ref_calib}
\end{table*}

\subsubsection{Verbalized Prompts with Reflection and Debate}
In this experiment, we evaluate the performance of advanced prompting techniques, including Reflection and Debate, using verbalized confidence estimation. As presented in Table \ref{tab:verbalized_rd}, Reflection achieves a significant improvement in AUPRC $(0.918)$ and AUROC $(0.687)$ compared to the log probability baseline, demonstrating its ability to generalize across datasets. Debate-based prompting slightly outperforms Reflection on AUROC $(0.692)$ and achieves the highest AUPRC $(0.927)$. These findings validate that advanced prompting strategies improve model performance. Notably, the improved ECE observed for Reflection and Debate can be attributed to the inherent class imbalance in the dataset and the model’s tendency to over-predict certain labels. This clustering of confidence scores within a narrow range positively impacts calibration metrics like ECE, underscoring the need for multi-metric evaluation.

\subsubsection{MCS with Reflection and Debate}
In this experiment, we evaluate the integration of Reflection and Debate into the MCS framework for confidence estimation. Table \ref{tab:mcs_reflection_debate} provides a comparative analysis of MCS with and without enhanced prompts. Key observations are:
\textbf{1) Incremental improvements:}  MCS-R achieves notable improvements in AUROC ($0.697$ vs. $0.591$) and AUPRC ($0.925$ vs. $0.899$) compared to the standalone MCS method.
\textbf{2) Reflection vs. Debate:} Reflection slightly outperforms Debate in AUPRC ($0.925$ vs. $0.924$) but lags in AUROC ($0.697$ vs. $0.7$). This contrast suggests complementary strengths between the two prompting strategies.
\textbf{3) Calibration limitations:} Despite improved discriminative performance, ECE remains higher for both Reflection ($0.301$) and Debate ($0.322$) compared to MCS, indicating the need for post-processing techniques like calibration.

\begin{figure*}[htbp]
    \centering
    \includegraphics[width=0.9\linewidth]{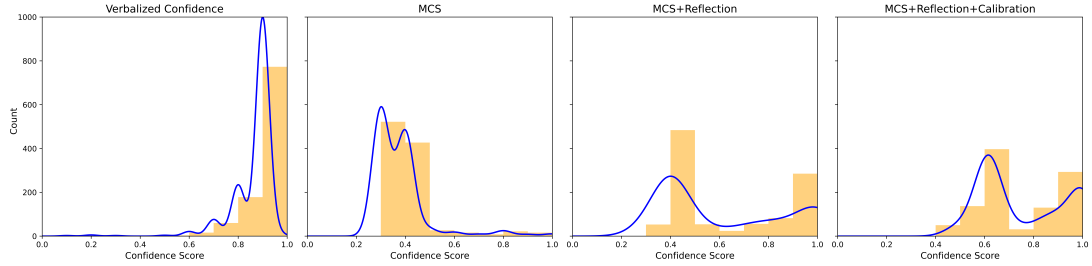} 
    
    \caption{Confidence score distribution across different methods: VC, MCS, MCS, MCS-R, and MCS-RC. The histogram illustrates the frequency of confidence scores, while the blue line represents the kernel density estimation. The progression from VC to MCS-RC demonstrates improved confidence score distribution, with reduced overconfidence and better alignment with model uncertainty}
    \label{fig:conf_dist}
\end{figure*}

\begin{table}[t]
\centering
\caption{Performance Comparison on T-REx Dataset}
\begin{tabular}{lc}
\hline
\textbf{Method} & \textbf{AUPRC} \\
\hline
Verbalized Confidence & 0.700 \\
Log Prob & 0.709 \\
Trained Probe (SOTA) & \textbf{0.910} \\
Verbalized Debate (Proposed) & 0.887 \\
Verbalized Reflection (Proposed) & \underline{0.890} \\
\hline
\end{tabular}
\label{tab:trex}
\end{table}

\subsubsection{MCS with Enhanced Prompts and Calibration}
This experiment incorporates distance-aware calibration into the MCS framework enhanced by Reflection and Debate. Calibration aligns confidence scores with ordinal relationships, mitigating overconfidence and aligning predictions with the underlying structure of the label set. Table \ref{tab:mcs_ref_calib} presents the results for calibrated and non-calibrated methods. Findings are: 
\textbf{Effectiveness of calibration:} Calibration significantly enhances AUROC (from $0.697$ to $0.739$) and AUPRC (from $0.925$ to $0.935$) for MCS+Reflection. Similar gains are observed for MCS+Debate, underscoring the utility of calibration.
\textbf{Better calibration metrics:}
Applying calibration reduces ECE by $41.9\%$ for Reflection and $40.2\%$ for Debate, ensuring confidence scores are better aligned with correctness. Similarly, Brier Score improves substantially, dropping by $51.2\%$ for Reflection and $49.9\%$ for Debate.
\textbf{Best-performing method:} The proposed method, MCS-RC, achieves the highest performance across all metrics, combining effective calibration and reasoning-driven confidence estimation to improve both calibration and discriminative capabilities.

\section{Ablation Studies}
\subsection{Generalizing to Diverse Dataset Type}

\textbf{Objective and Hypothesis} We investigate whether our proposed prompting techniques, Reflection and Debate, generalize effectively to factual classification tasks. Specifically, we compare these methods against state-of-the-art approaches, including VC, TP, and LP, on the $13.6K$ examples of Llama-T-REx dataset, as constructed in Section \ref{sec:datasets}. Our hypothesis is that advanced prompting techniques provide a robust alternative to fine-tuning, offering comparable or superior performance. Our findings from Table \ref{tab:trex} are as follows:
\textbf{1) Advanced prompting matches fine-tuning performance:} Reflection achieves an AUPRC of $0.89$, closely matching the Trained Probe method ($0.91$), which requires fine-tuning. This supports our hypothesis that advanced prompts provide a robust alternative to fine-tuning for confidence scoring.
Debate also performs strongly, with an AUPRC of $0.887$, demonstrating the consistency of advanced prompts.
\textbf{2) Significant improvement over vanilla prompts:} Both Reflection ($0.89$) and Debate ($0.887$) significantly outperform vanilla verbalized prompts ($0.70$) and Log Probability ($0.709$), validating the effectiveness of reasoning-driven confidence estimation.
\textbf{3) Generalizability across dataset types:} The strong performance of Reflection and Debate on Llama-T-REx, a factual classification dataset, demonstrates the generalizability of our advanced prompting techniques across diverse dataset types.

\subsection{Effect on Confidence Distribution}
We study the impact of different components of our proposed method on the distribution of confidence scores for \textit{helpfulness}, as illustrated in Figure \ref{fig:conf_dist}. The four subplots correspond to the distributions for VC, MCS, MCS-R, and MCS-RC.
The confidence distribution for VC methods is heavily right-skewed, reflecting  overconfidence. Applying the MCS method results in a more balanced distribution, improving the differentiation between confident and less confident predictions. MCS-R further smooths the distribution by allowing the model to refine its confidence estimates through a reconsideration of its initial reasoning, reducing extreme scores and improving alignment with correctness. Finally, integrating Distance-Aware Calibration with MCS-R, i.e. MCS-RC produces the most balanced distribution by redistributing confidence scores based on ordinal label relationships, effectively mitigating overconfidence and ensuring well-calibrated predictions.

\section{Conclusion}

We introduce a novel framework (MCS-RC) that integrates Maximum Confidence Selection, Reflection-based prompting, and Distance-Aware Calibration. Our experiments across multi-turn conversation and factual classification datasets show that Reflection and Debate prompting outperform traditional verbalized techniques, matching fine-tuned approaches while maintaining zero-shot flexibility. The framework improves AUPRC and AUROC metrics through two key mechanisms: Reflection enhances reasoning-driven confidence estimation, while Distance-Aware Calibration mitigates overconfidence by considering ordinal label relationships. Notably, the MCS-RC framework achieves these improvements without adding computational overhead, ensuring scalability for real-world applications. Confidence distribution analysis further highlights its ability to produce well-calibrated and interpretable scores, addressing critical challenges in trust and reliability for LLM-based systems.

\section{Limitations}
The reliance on distance-aware calibration assumes that the label space has a well-defined ordinal structure, which may not generalize to tasks with nominal or hierarchical labels. Although the zero-shot nature of our framework ensures computational efficiency and scalability, it may limit performance in scenarios where fine-tuning or task-specific adjustments could further enhance confidence estimation. Additionally, our experiments primarily focus on conversational and fact-based classification tasks, leaving open questions about the framework’s effectiveness in other domains, such as vision-language models or multi-modal tasks. These limitations underscore important directions for future work, including extending the framework to non-ordinal tasks, exploring other functions for calibration to replace distance-aware function, and validating its robustness across a wider range of applications and modalities.

\bibliography{custom}



\clearpage

\appendix

\section{Evaluation Metrics}
\label{app:eval_metrics}

These metrics collectively provide a comprehensive evaluation of our framework's performance, addressing both discrimination ability and calibration quality.

\subsection{Expected Calibration Error (ECE)}  ECE (Expected Calibration Error) is a measure used to evaluate the accuracy of a model's confidence predictions. Ideally, a model's confidence should accurately represent the actual likelihood that its predictions are correct. The ECE assesses how well the predicted probabilities match the true outcomes by grouping these probabilities into specified intervals or bins and then evaluating the average discrepancies within those bins.
The Expected Calibration Error is calculated by: 1. Dividing the range of predicted probabilities into a set number of bins or intervals.
2. For each bin, calculating the absolute difference between the mean predicted probability (confidence) and the actual accuracy.
3. Computing the weighted average of these differences across all bins to obtain the ECE.

Formula and Explanation:

\begin{equation}
    \label{eq:ece}
    ECE = \sum_{m=1}^{M} \left(\frac{|B_m|}{n}\right) |acc(B_m) - conf(B_m)|
\end{equation}

where $M$ is the total number of bins. $B_m$ represents the set of samples within the $m^{th}$  bin. $n$ is the total number of samples. $acc(B_m)$ is the accuracy within the $m^{th}$ bin, defined as the proportion of correct predictions. $conf(B_m)$ is the average predicted probability (confidence) within the $m^{th}$ bin.

\subsection{Brier Score} The Brier score measures the mean squared difference between the predicted probability assigned to the possible outcomes and the actual outcome. It evaluates how well-calibrated the predicted probabilities are. The Brier score measures the accuracy and calibration of probabilistic predictions. A Brier score of 0 indicates a perfect model.

\begin{equation}
    \text{BS} = \frac{1}{N} \sum_{t=1}^{N} (f_t - o_t)^2
\end{equation}

where:
\begin{itemize}
    \item $f_t$ is the predicted probability
    \item $o_t$ is the actual outcome (0 or 1)
    \item $N$ is the number of predictions
\end{itemize}

\subsection{AUPRC} AUPRC, or the Area Under the Precision-Recall Curve, evaluates the performance of a model by considering the trade-off between precision and recall at various confidence thresholds. It is particularly well-suited for imbalanced datasets where one class significantly outweighs the other.

\begin{equation}
    \label{eq:auprc}
    AUPRC = \sum_{n=1}^{N} (R_n - R_{n-1}) \cdot P_n
\end{equation}

$N$: The number of points in the precision-recall curve, $P_n$: The precision at the $n^{th}$ threshold, $R_n$: The recall at the $n^{th}$  threshold, $R_{n-1}$: The recall at the previous threshold, $R_n - R_{n-1}$: The change in recall between consecutive thresholds.

\subsection{AUROC}

AUROC, or the Area Under the Receiver Operating Characteristic Curve operates by defining a function $R(x,y)$, which is set to $1$ if the model's predicted answer $y$ for an input $x$ is correct, and $0$ otherwise. Concurrently, $C(x)$ denotes the model’s confidence in its prediction for $x$, ranging between $0$ and $1$.

Formulas and Explanations:
True Positive Rate (TPR): This rate is calculated at a specific confidence threshold $t$ and represents the proportion of correctly predicted samples that have a confidence level equal to or greater than 
$t$. The formula for TPR is given by:

\begin{equation}
    \label{eq:auroc1}
    TPR(t) = \frac{\sum [R(x, y(x)) \cdot I(C(x) \geq t)]}{\sum [R(x, y(x))]}
\end{equation}

Here, $I$ is an indicator function that is $1$ if $C(x)>=t$ and $0$ otherwise.

False Positive Rate (FPR): FPR measures the ratio of incorrectly predicted samples that have a confidence level of t or higher. The formula for calculating FPR is:

\begin{equation}
    \label{eq:auroc2}
    FPR(t) = \frac{\sum [(1 - R(x, y(x))) \cdot I(C(x) \geq t)]}{\sum [1 - R(x, y(x))]}
\end{equation}
This calculation also employs the indicator function $I$ similar to the TPR formula.

To construct the ROC curve, TPR and FPR values are plotted for various thresholds $t$. The AUROC is then determined by calculating the area under this curve. A higher AUROC value (close to $1$) signifies better discriminative ability of the classifier, indicating it is capable of distinguishing between the classes effectively, while a lower value (close to $0$) suggests poor performance.

\section{Verbalized Prompting Methods}
\label{app:verbalized_methods}
\subsection{Chain of Thoughts}

Chain of Thought prompting can be used to improve the explanation and justification behind each model's decision. By breaking down its reasoning into smaller steps, the model can not only provide a final prediction but also explain the intermediate logic that leads to this prediction, making the confidence score more interpretable.


\subsection{Few-Shot Learning}
By providing a few examples of correct and incorrect predictions, the model can better gauge its own performance and provide a more accurate probability score for its predictions.

\subsection{Self-Consistency}
Self-Consistency can be employed to estimate the confidence score by running multiple reasoning paths and evaluating the consensus across them. If most paths lead to the same prediction, the confidence score should be high. Conversely, if the model generates diverse or conflicting outputs, the confidence score would be lower, providing a probabilistic assessment of the prediction's reliability.

\subsection{Debate-Style Prompting}
Debate-style prompting can be integrated into confidence scoring by having the model argue for and against its predicted label. If the arguments supporting the predicted label consistently outweigh the counterarguments, the model can assign a higher confidence score to its prediction. This method allows the model to internally challenge its decisions, refining the accuracy of its confidence estimation.

\subsection{Reflection Prompting}
Incorporating Reflection Prompting would involve the model self-assessing its initial prediction and offering an explanation of potential errors. By reflecting on possible mistakes and refining its answer, the model can provide a more accurate and justified confidence score. Reflection increases the model's ability to adjust its confidence level after a self-evaluation, improving overall reliability in probabilistic outputs.









\section{Prompts for Proposed methods}
\label{app:prompt_templates}

\begin{figure*}[t]
\begin{tcolorbox}[title=Prompt for Verbalized Confidence]
\label{app:verbalized}
\begin{verbatim}
Provide your confidence level (on a scale of 0.0 to 1.0) that the following
statement is correct.
The statement is: {statement} 
Confidence level:
\end{verbatim}
\end{tcolorbox}
\end{figure*}

\begin{figure*}[t]
\begin{tcolorbox}[title=Prompt for MCS method]
\begin{verbatim}
You are an AI judge tasked with the assessment of the quality of interaction
between a user and a conversation agent. You are presented with a single-turn
interaction between the USER and AGENT, which contains a USER utterance\/request
and a conversational AGENT response.

### Metric: {metric}
### Score Rubrics: {rubrics}

### Your tasks:
Your task is to assign a probability of likelihood of each class in scoring
rubric being correct.
Estimate the probability (between 0 and 1) that each label is correct.

### Please output the following:
(Score is a class from the Score Rubrics. It can have only the actual class
label such as 1,2,3,4,5)
**score_n:** Probability of nth score in Score Rubrics being correct
---
USER: {user}
AGENT: {agent}
\end{verbatim}
\end{tcolorbox}
\end{figure*}

\end{document}